\documentclass[conference]{IEEEtran}
\IEEEoverridecommandlockouts
\usepackage{cite}
\usepackage{amsmath,amssymb,amsfonts}
\usepackage{algorithmic}
\usepackage{graphicx}
\usepackage{textcomp}
\usepackage{xcolor}
\usepackage{glossaries-extra}
\usepackage{hyperref}
\usepackage[capitalise]{cleveref}
\usepackage{glossaries-extra}
\usepackage{booktabs}
\usepackage{siunitx, physics}
\usepackage{gensymb}
\usepackage{pifont}% http://ctan.org/pkg/pifont
\setabbreviationstyle[acronym]{long-short}
\glssetcategoryattribute{acronym}{nohyperfirst}{true}

\makeglossaries

\newacronym{cnn}{CNN}{convolutional neural network}
\newacronym{ddt}{DDT}{dynamic driving task}
\newacronym{dgt}{DGT}{dynamic ground truth}
\newacronym{ptz}{PTZ}{panning, rotation, and zoom}
\newacronym{dsnet}{DSNet}{Dual-Siamese Network}
\newacronym{sift}{SIFT}{scale-invariant feature descriptors}
\newacronym{usm}{USM}{Unified Spherical Model}
\newacronym{sfm}{SfM}{Structure-from-Motion}
\newacronym{icp}{ICP}{Iterative Closest Point}
\newacronym{calica}{CaLiCaNet}{Camera Lidar Calibration Network}

\begin{document}

\title{End-to-End Lidar-Camera Self-Calibration for Autonomous Vehicles\\

\thanks{We gratefully acknowledge that this work has been supported by the Bavarian State Ministry of Economic Affairs, Regional Development, and Energy (StMWi) under the EmPer project (grant no. DIK 0179/01).}
}

%\author{
%    \IEEEauthorblockN{%
%        AuthorA\IEEEauthorrefmark{1}\,\IEEEauthorrefmark{2}  \qquad 
%        AuthorB\IEEEauthorrefmark{1}  \qquad AuthorC \IEEEauthorrefmark{1}
%    } \\  \IEEEauthorblockA{%
%        \IEEEauthorrefmark{1}Friedrich-Alexander-Universität Erlangen-Nürnberg\\
%           Multimedia Communications and Signal Processing\\
%            Erlangen, Germany \\
%            {firstname.lastname@fau.de}
%    }
%\hfill
%\IEEEauthorblockA{%
%    \IEEEauthorrefmark{1}AVL Software and Functions GmbH\\
%    Product Segment Autonomous Driving\\
%    Regensburg, Germany 
%}
% 
%}%end author

\author{
   \IEEEauthorblockN{%
               Arya Rachman\IEEEauthorrefmark{1}\,\IEEEauthorrefmark{2}  \qquad 
               Jürgen Seiler\IEEEauthorrefmark{1}  \qquad André Kaup\IEEEauthorrefmark{1}
           }\\
    \IEEEauthorblockA{\begin{tabular}{cc}
           \IEEEauthorrefmark{1}Multimedia Communications and Signal Processing & \IEEEauthorrefmark{2}Product Segment Autonomous Driving\\
            Friedrich-Alexander-Universität Erlangen-Nürnberg &  AVL Software and Functions, GmbH \\
            Erlangen, Germany & Regensburg, Germany \\ 
            Email: \{firstname.lastname\}@fau.de & 
    \end{tabular}}
}

%
% {Arya Rachman $^{\dagger \star}$, Jürgen Seiler$^{\dagger}$, and André Kaup$^{\dagger}$\thanks{We gratefully acknowledge that this work has been supported by the Bavarian State Ministry of Economic Affairs, Regional Development, and Energy (StMWi) under the EmPer project (grant no. DIK 0179/01).}}
%%{$^{\dagger}$Friedrich-Alexander-Universität Erlangen-Nürnberg\\
%%    Multimedia Communications and Signal Processing\\
%%    Erlangen, Germany}
%%{}
%{
%    $^{\star}$AVL Software and Functions GmbH\\
%    Product Segment Autonomous Driving\\
%    Regensburg, Germany}
%

\maketitle

\begin{abstract}
Autonomous vehicles are equipped with a multi-modal sensor setup to enable the car to drive safely. The \textit{initial} calibration of such perception sensors is a highly matured topic and is routinely done in an automated factory environment. However, an intriguing question arises on how to maintain the calibration quality throughout the vehicle's operating duration. Another challenge is to calibrate multiple sensors jointly to ensure no propagation of systemic errors. In this paper, we propose \gls{calica}, an end-to-end deep self-calibration network which addresses the automatic calibration problem for pinhole camera and Lidar. We jointly predict the camera intrinsic parameters (focal length and distortion) as well as Lidar-Camera extrinsic parameters (rotation and translation), by regressing feature correlation between the camera image and the Lidar point cloud. The network is arranged in a Siamese-twin structure to constrain the network features learning to a mutually shared feature in both point cloud and camera (Lidar-camera constraint). Evaluation using KITTI datasets shows that we achieve 0.154$\degree$ and 0.059 $\si\metre$ accuracy with a reprojection error of 0.028 pixel with a single-pass inference. We also provide an ablative study of how our end-to-end learning architecture offers lower terminal loss (21\% decrease in rotation loss) compared to isolated calibration.

%KITTI Datasets, as well as data taken from a Dynamic Ground Truth system are used for evaluation.
\end{abstract}


\section{Introduction}
Modern vehicles are equipped with multi-modal sensors, enabling perception algorithms to better understand the environment. Compared to classical camera-only perception, adding depth-capable sensors such as Radar and Lidar enables unambiguous recognition in a 3D world. Therefore, a camera and Lidar combination is the most used in autonomous vehicle setup, often with multiple instances of the same modality (e.g., 2$\times$ Lidar plus 3$\times$ camera).

The use of such multi-sensor setups comes with the need for cascaded calibration: first, each sensor needs to be both calibrated intrinsically relative to its internals, and after that, it needs to be extrinsically calibrated to other sensors. Both calibrations are prerequisites for sensor fusion, a building block of vehicle perception. In order to perceive the environment consistently, the vehicle has to use sensor measurements in a common coordinate system.

The quality of intrinsic calibration generally impacts extrinsic calibration. Minor errors in the intrinsic calibration may affect the extrinsic calibration's accuracy. To tackle this problem, jointly-optimized intrinsic and extrinsic calibration is proposed in\cite{Hillemann2017}\cite{Kodaira2022}. The approach offers better performance compared to isolated calibration. However, it requires a static target which is not applicable when a vehicle is in use. For this case, a targetless calibration method is needed.

The classical model-based self-calibration based on \gls{sfm}\cite{Nagy2020}, optical flow\cite{Kodaira2022}, and odometry\cite{Heng2013} also exist. However, in order to accommodate a broad operating domain and avoid heuristically tuned algorithms, deep-learning-based approaches have been increasingly used for self-calibration (see \cite{Bogdan2018,Lv2021,Zhang2020}). Notwithstanding, the comparison between model-based vs. deep-learning-based approaches remains difficult since there is little-to-no back-to-back comparison, and each approach has a different operating domain, not necessarily automotive. Finally, the deep learning approach primarily relies on supervised networks: this means there is a need for a sufficient amount of labelled, high quality data for the network to properly generalize.

Our previous work\cite{Rachman2022} introduced a self-calibration pipeline to perform intrinsic camera calibration using deep neural networks and a back-to-back comparison against a target-based method. In this paper, we extend the pipeline with a novel end-to-end learning network called \gls{calica}, aiming to calibrate a camera intrinsically from scratch and, simultaneously, extrinsically to a Lidar. This is a distinction from the majority of existing Lidar-Camera calibration methods that assume perfect camera intrinsic parameters to be available. 

%We recognize in addition to novelty, it is important as well to demonstrate feasibility of our proposal, to this effect
To measure the impact of the proposed architecture, we use evaluation metrics that are directly comparable to conventional pattern-based calibration and benchmark our proposal to a dataset that reflects the target domain: KITTI Datasets\cite{Geiger2013}. We also include an ablation study treating the intrinsic and extrinsic networks as sub-networks. The aim is to understand  how end-to-end networks perform better than isolated networks for Lidar-Camera calibration.
%as well as our own data taken from a Dynamic Ground Truth system\cite{Tihanyi2021} are used for evaluation.  
\newpage
To summarize, our contributions consist of the following:

\begin{enumerate}
    \item We propose end-to-end \gls{calica}, which to the best of our knowledge is the first application of end-to-end learning for the purpose of sensor calibration.
    \item We evaluate our approach using KITTI Dataset with the specific metrics directly comparable to the conventional calibration method. This facilitates an informed decision to switch from the state-of-the-practice to the state-of-the-art calibration approach.
    \item We propose a streamlined data collection and label generation pipeline for a self-calibrating automotive Lidar-Camera system.
\end{enumerate}
% We will also publish select dataset collected with the proposed pipeline along with the checkerboard calibration dataset. To the best of our knowledge only KITTI Dataset provides a checkerboard recording for independent verification of calibration quality. The select datasets are to made publicly to encourage future research in this topic.

\section{Related Works}
Lidar-Camera extrinsic calibration is a well-visited topic. Typically, as a first step, a camera is intrinsically calibrated using a geometric pattern like a checkerboard\cite{Zhang2000}. Then, its extrinsic calibration to Lidar is done by matching 3D features from the Lidar point cloud to the camera equivalent (i.e., in the image plane). The 3D features from the camera image can be generated from the 3D checkerboard plane or 3D reconstruction method, such as \gls{sfm}, given the intrinsic parameters.

In practice, the camera intrinsic calibration and Lidar-Camera extrinsic calibration are done (1) sequentially and (2) performed using a specialized pattern before the sensor system is used for intended operation (i.e., offline). Calibrating the camera first, followed by the Lidar-Camera, poses the risk of a systemic error propagating to extrinsic calibration when intrinsic parameters are not sufficiently accurate. Offline calibration also does not consider mechanical shifts occurring during the sensor system operation. The shift is especially relevant for an automotive system, requiring modern vehicles equipped with an array of perception sensors to be periodically recalibrated. 

To address the first challenge, we refer to the formulation of joint Lidar-Camera calibration as proposed in \cite{Tu2022,Ou2022,Hillemann2017}. The general idea is to reach a more robust and globally optimal solution bounded by physical measurements of multi-modal sensors. With regard to the second challenge, a global optimization is ideally done without the need for a pattern-like target and uses a natural environment to enable the possibility of online calibration. The  \gls{sfm}-based approach\cite{Tu2022} comes close to our needs: natural features are detected using \gls{sfm} and used as inputs to the Bundle Adjustment algorithm to predict intrinsic parameters. Lastly, \gls{icp} is used for extrinsic calibration. 

Notwithstanding, feature detection and matching using an explicit model require a stable and feature-rich environment and, therefore, generally is not very robust when applied to online calibration. Furthermore, in \cite{Tu2022}, the evaluation using KITTI dataset is limited to road scenarios (e.g., stable environment), similar limitation can also be found in \cite{Nagy2020}, where it relies on pillar-like structures for the calibration to work. The model-based feature extractor, such as SfM or odometry, typically requires heuristics parameter tuning that needs to be changed once the operating domain (e.g., camera type or environment) is shifted.

Acknowledging these general difficulties, deep-learning-based perception sensor calibration is proposed, as seen in \cite{Bogdan2018,Lv2021,Zhang2020}. The main idea is to leverage well-validated and pre-trained image detectors (e.g., ImageNet, Inception, ResNet), which are conventionally used for image classification and object detection, as an environmental feature extractor.

For automotive purposes, DeepPTZ\cite{Zhang2020} is able to calibrate vehicle cameras with the extension we provided in \cite{Rachman2022}. We also consider LCCNet\cite{Lv2021} and CFNet\cite{Lv2021a}, which evaluate their method directly on KITTI Dataset for Lidar-Camera calibration. Still, these deep-learning approaches have yet to address the need to jointly calibrate camera intrinsic and camera extrinsic.

Unlike the classical model-based online calibration, we believe there is no learning-based solution yet that performs Lidar-Camera calibration end-to-end, the implicit assumption being that perfect camera intrinsic is always available and always rigid notwithstanding with a vehicle dynamics (e.g., vibration) influencing sensor and outdoor environment (e.g., temperature) affecting sensor mechanical setup. 

Finally, to the best of our knowledge, end-to-end learnings were widely investigated, albeit within a limited domain (e.g., self-driving\cite{Tampuu2022} or text recognition\cite{Shi2017}). Therefore, the behaviour of end-to-end network design is only little visited in the field of vehicle perception, especially sensor calibration.
\begin{figure*}[!t]
    \centering
    \includegraphics[width=0.95\linewidth]{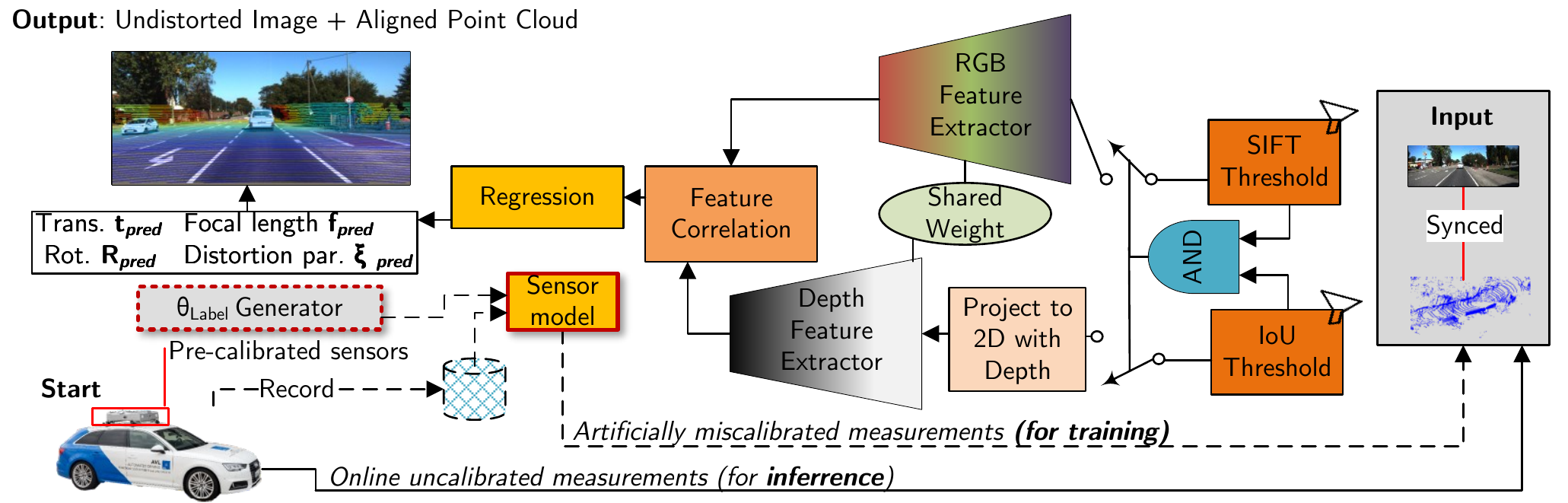}
    \caption[End-to-end Self-Calibration Pipeline]{End-to-end Lidar-Camera Self-Calibration Pipeline with \gls{calica}. The parts denoted by the dashed line are only used during network training. }
    \label{fig:arch}
\end{figure*}
\section{Self-Calibration Problem}
\subsection{Problem Formulation}
\label{subsec:models}
For the camera, we consider the pinhole model following \cite{Hartley:2003:MVG:861369}, by which lens distortion is modelled using the \gls{usm}. The camera intrinsic calibration is formulated as recovering intrinsic parameters $\vb{\textit{K}}_i$, and distortion coefficient $\xi_i$ from $i$ input images.
 % can be extended as well to include tangential and radial distortion as described in \cite{}, but these bring disadvantage of losing the closed-loop formulation as polynomial model can be only inverted with approximation and increases the number needed by the parameters seen by the network. 

Meanwhile, the Lidar extrinsic calibration is described by a rigid 3D transform consisting of rotation $\vb{\textit{R}}_i$  and translation $\vb{\textit{t}}_i$. For each image $i$ and point cloud $(X,Y,Z)_i$, a point in the 3D world can be projected using matrix $\vb{\textit{K}}$  to camera pixel coordinate $(u,v)$, assuming an undistorted image and no scaling is needed:

\begin{equation}
    \left[\begin{array}{l}
        u \\
        v \\
        1
    \end{array}\right]=\vb{\textit{K}}[\vb{\textit{R}} \vb{\textit{t}}]\left[\begin{array}{l}
        X \\
        Y \\
        Z \\
        1
    \end{array}\right]
\label{eq:extrinsic}
\end{equation}

%\textbf{Note: } Here, it is evident that inaccuracy in intrinsic parameters $\vb{K}$ will affect the extrinsic calibration, as they have affine relationship.
%$\vb{K},  

When the intrinsic camera parameters and Lidar-Camera parameters estimation are treated within a unified optimization problem, we get a better chance of coming to a globally optimal solution\cite{Yan2022}. The advantage is especially relevant since joint optimization provides the physical constraints recognized by both Lidar and the camera.

Based on the above formulation, we delineate our goal of end-to-end Lidar-Camera self-calibration to the acquisition of calibration values  $\theta = \{ {f}, \vb{\xi}, \vb{\textit{R}}, \vb{\textit{t}} \} $ given sets of image and 3D point cloud pair from natural scenes. The image sensor is assumed to be square (focal length $f$ is symmetrical along the  $x$- and $y$-axis), and the Lidar scan is assumed to be ego-motion-compensated. The constraint is that the measurement does not need to have an explicit geometric pattern nor be taken in a specific environment with certain structure features (e.g., pole or building). Considering the difficulty of modelling such an environment, we defer to a learning-based approach.
\subsection{Deep Learning by Driving: Generating Training Data}
Deep learning approaches for sensor calibration are largely supervised and thus require labelled ground truth. We refer to the learning-by-driving label generation strategy previously proposed in {\cite{Rachman2022}. Our training label generation strategy does not need human annotation nor additional infrastructure not already existing in a modern car and automotive supply chain. First, we pre-calibrate a pair of camera and Lidar sensor with a checkerboard-based, controlled environment setup to get $\theta_0$. We consider $\theta_0$ as ground truth calibration values. We then collect the sensor measurements while the car is driving with the target calibration sensors. 

The resulting measurements become a dataset of time-series calibrated frames (point cloud and image). Analogous to augmentation, \textit{realistic} (i.e., based on known calibration deviation over a long-term driving period) $\theta_{label}$ are generated together along with sets of distorted images and miscalibrated point clouds, as seen in \cref{fig:arch}.

\section{End-to-End Design}\label{AA}
\subsection{Isolated Networks}
In this work, we adopt networks with public reference implementation as benchmarks and inspirations for our end-to-end design. Following (\ref{eq:extrinsic}), the output of an intrinsic calibration network can be chained directly to an extrinsic calibration network. In this case, the prediction and training of each intrinsic and extrinsic network happen in isolation.

For the intrinsic networks, we adopt some part of DSNet\cite{Zhang2020}, a Siamese-network designed to predict camera intrinsics calibration $f$ and $\xi$ with inputs of two correlated images from different viewpoints. It was originally designed to work with panoptic datasets and \gls{ptz} camera, but we have extended it in \cite{Rachman2022} to work with vehicle cameras as the target domain. As for extrinsic calibration, inspired by LCCNet\cite{Lv2021}, we use a volume-cost-based network that correlates the volume cost shared by the image and the projected point cloud depth image. Both isolated networks rely on features correlation to regress $\theta$. 
%\begin{figure*}[!ht]
%    \centering
%    \includegraphics[width=0.8\linewidth]{fig/end2end_vis}
%    \caption[End-to-end Self-Calibration]{End-to-End Self-Calibration Input/Output.}
%    \label{fig:vis}
%\end{figure*}

\begin{figure*}[!ht]
    \centering
    \includegraphics[width=1.0\linewidth]{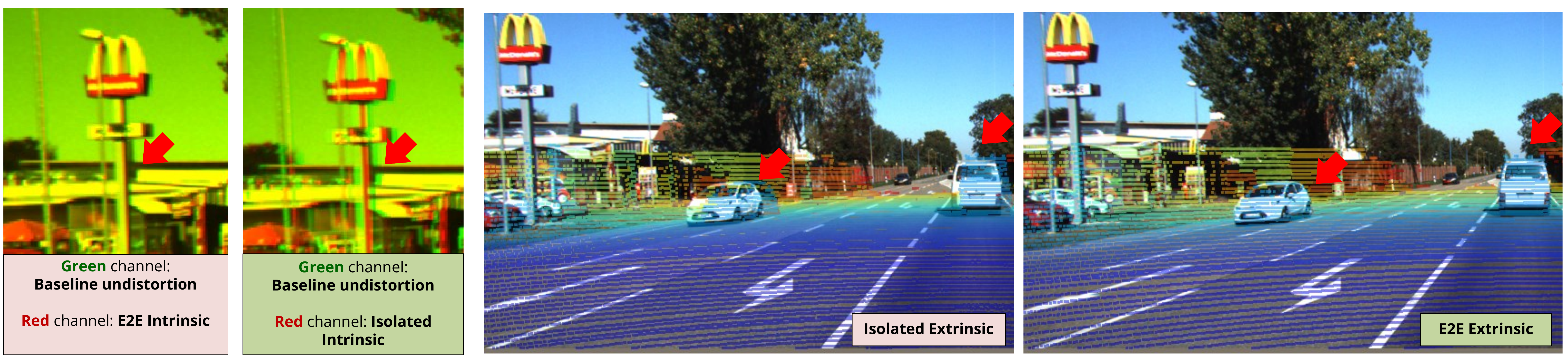}
    \caption[End-to-end Self-Calibration]{End-to-End vs Isolated Self-Calibration. \textbf{Intrinsic}: red/green halos indicate pixel-wise differences. \textbf{Extrinsic}: projected point cloud to an image representing alignment quality. Red arrows point to misalignment regions of interest.}
    \label{fig:vi_e2e_vs_iso}
\end{figure*}

\subsection{CalicaNet}
\gls{calica} (Camera Lidar Calibration Network) aims to enable effective end-to-end learning by building upon feature extraction networks to associate features from two correlated sensor measurements and regress them into calibration parameters.

Recall that the DSNet feature map is extracted based on two-correlated monocular images to enforce a bi-directionality constraint. While this has been shown to help the network to focus on relevant features, the features from monocular cameras are inherently two-dimensional. Making the network susceptible to the famous "Wile E. Coyote's tunnel" (a tunnel imagery being painted on a wall to induce illusion of depth).  We address the problem by enforcing the constraint from two different sensor modalities: Lidar and camera. Compared to a single modality approach, environmental attenuations specific to a camera (e.g., low light or fake depth) are mitigated by a Lidar, and vice versa (e.g., dark object or sparsity).

On the other hand, LCCNet relies on the assumption that the input image is perfectly undistorted and that the camera intrinsic matrix $\vb{K}$ is sufficiently accurate. If the lens undistortion is inaccurate, for example, the Lidar-Camera extrinsic prediction can be overfitted to the centre region of the image. More importantly, LCCNet requires multi-pass inferences to achieve  sufficient accuracy for autonomous driving. With our \gls{calica}, we mitigate this problem by explicitly incorporating intrinsic parameters in the loss function and implementing the Lidar-Camera constraint by means of weight sharing. 
Further details of \gls{calica}} can be found in \cref{fig:arch} as well as in the following subsections:

\subsubsection{\textbf{Pre-processing: ensuring consistent input}}
We discard featureless images (e.g., frames dominated by sky and rural, empty road) by computing SIFT descriptors on each frame and passing only frames with a certain threshold of key-points count (see \cite{Lowe2004} for details). Additionally, we also ensure the overlap with the Intersection of Union (IoU)) of $\geq 0.5$ between an RGB image and the corresponding Lidar 2D projection. These measures are intended so that the network sees more consistent input during training and inference.
\subsubsection{\textbf{Network Architecture}}
We use ResNet50\cite{He2016} as a feature extractor and implement the necessary fully connected layers for generating the correlation between inputs. Since we aim to use single-pass inference from natural scenes, we choose a deeper network block. In contrast, DSNet uses Inceptionv3, and LCCNet uses ResNet18. We also aim to accept an arbitrary resolution on the training images and point cloud, owing to the diversity of vehicle sensors; therefore, we place an adaptive average pooling before the fully connected layer. Furthermore, we extend the volume-cost-based correlation module from \cite{Lv2021} to work with a Siamese structure. Finally, we opt to use Parametrized ReLU (PReLU) for our network extractor activation function. The ReLU weight, therefore, becomes a trainable parameter. Our functional network architecture as part of the end-to-end pipeline is described in \cref{fig:arch}.

\subsubsection{\textbf{Loss Functions}}
The network minimizes the L2-smooth loss $L$ between regressed calibration values of $\theta_{pred}$ and that of label $\theta_{label}$. Except for $\vb{R}$ that is converted to equivalent quaternion representation $\vb{q}$, and therefore quaternion distance is treated as a loss. For vehicle perception purposes, the misalignment in translation does not affect vehicle perception as much as rotation (refer to \cref{fig:vi_e2e_vs_iso} for an example of misalignment due to rotation). Therefore, we penalize the network more for predicting the wrong rotation. The training loss $L_T$ thus becomes:

\begin{equation}
    L_T=\lambda_f L_f + \lambda_\xi L_\xi + \lambda_q L_q + \lambda_t L_t 
\end{equation}

with $\lambda_q \ge \lambda_t $

%One of our contribution lies on the streamlining of Siamese network design into Lidar-Camera end-to-end calibration. In previous literature\cite{} the bidirectional constraint is achieved by training the intrinsic prediction network using two correlated input images from different camera perspectives. 

%2The common representation of Lidar and Camera for the purpose of Lidar-Camera extrinsic calibration is introduced in \cite{}. We follow LCCNet \cite{} design which project the 3D point cloud into a 2D image with a depth channel (Depth branch), which in turn can be matched to camera image (RGB branch). LCCNet requires multiple passes of training and inference to achieve good extrinsic calibration. In CalicaNet, we added siamese architecture between Depth and RGB branch to help the network extract relevant features by means of bidirectional constant. In our ablation study (\cref{}) we have shown that comparable accuracy in single pass inference is achieved. Finally, we recognize for autonomous driving purpose, time-constrained, online-capable calibration procedure is desired, we designed CaLiCaNet considering its fitness for online calibration pipeline.

\section{Evaluation}\label{AA}

We propose two evaluation strategies: the first is evaluating with KITTI Dataset to check the performance in a realistic situation. This serves as validation of our approach. The second strategy is an ablation study to examine closer how an end-to-end architecture enhances the network performance; this serves as verification of our network design. Such two-step approach is a necessary first step in developing safety-critical software running on a vehicle\cite{Zhang2020b}.

\subsection{Experimental Setup}
\label{subsec:exp_setup}
\subsubsection{\textbf{Dataset}}To evaluate the calibration accuracy, we use the  KITTI Dataset \cite{Geiger2013}, in which the RGB images captured by the right-side monocular camera are used. Ground-truth calibration values  $\theta_{0}$  from the KITTI dataset are generated as follows: we use the OpenCV Omnidirectional Calibration library \cite{opencv_library} to perform intrinsic calibration for each daily drive using the corresponding checkerboard recordings. Following the models stated in \cref{subsec:models}, we obtain $\theta_0$ and the corresponding baseline reprojection root mean square (RMS) error $\epsilon_{0}$ \cite{Hartley:2003:MVG:861369}.

\subsubsection{\textbf{Label Generation and Training}}The first day of the KITTI drive (2011-09-26) is used as training labels, while the remaining four days are used for evaluation. We generated the labels by setting the labels' value to $\theta_{0}$ plus a variable deviation (i.e., the miscalibration). The label generation refers to physical changes on the sensors induced by mechanical and environmental factors during everyday driving and is set to  $\theta_{label} = \{ {f} \pm 100 \si{px}, \vb{\xi} \pm 0.48, \vb{R} \pm 2.0 \degree, \vb{t} \pm 0.2 \si{m} \}$. The distorted image and misaligned point cloud pairs are generated according to $\theta_{0}$. In total, we generated approximately \SI{120000} labels ($\theta_{label}$, image and point cloud) split into 80:20 ratios for training and validation datasets. Adam Optimizer is used with an initial learning rate of \num{3e-3}. The batch size used was 60 with maximum epochs of 300.

\subsubsection{\textbf{Evaluation Metrics}}

For calculating the intrinsic calibration accuracy, we use KITTI's checkerboard which contain $i$ corners $\left\{p_{1j}, p_{2j}, \cdots, p_{ij}\right\} \in \mathbb{R}^3$ from $j$-th checkerboard poses $\vb{R} \vb{t}$. The purpose is to calculate the Root Mean Squared Error (RMSE) of the reprojection $\vb{\epsilon}$ in pixels, given as:
\begin{equation}
    \epsilon = \sqrt{\frac{1}{N}\sum_{i=1} \sum_{j=1} \left\|\vb{K}[\vb{R} \vb{t}]p_{ij}-x_{i j}\right\|_2}
\end{equation}

Note that this reprojection error value obtained based on $\theta_{0}$ is considered as the \textit{baseline} reprojection error $\epsilon_{0}$. When our network infers $\theta_{pred}$, the detected corners and checkerboard poses are reused to calculate inference reprojection error $\epsilon_{pred}$. We consider $\epsilon_{pred}$ to be more representative of calibration quality relative to the comparison of the intrinsic part of $\theta_{0}$ to $\theta_{pred}$. We refer to \cite{Hartley:2003:MVG:861369} for more in-depth reasoning.

On the other hand, for extrinsic calibration, for the sake of comparability, we opt to follow the convention in prior works \cite{Lv2021,Nguyen2022,Tu2022} by directly comparing the extrinsic part of $\theta_{0}$ to that of $\theta_{pred}$, that is the mean error of rotation and translation. In this case, the baseline extrinsic error is assumed to be zero.
%% \footnote{\url{https://docs.opencv.org/4.x/d3/ddc/group\_\_ccalib.html}}
%
%%
% We divide the KITTI dataset by the date the recordings were taken. Note that, for each day, checkerboard frames intended for calibration are also provided, and thus it can be assumed that the calibration is different for each date. 
% Due to minimal to no weather variation in the KITTI dataset, we also complement the experiment with three additional recordings taken in Roding using AVL's \gls{dgt}\cite{Tihanyi2021} (refer to  \cref{fig:calibpipeline} to see the physical setup). Past recordings in several environments (highway, city, interurban, etc.) with differing weather (e.g., sunny and rain) are used for training. New recordings were taken for evaluation. Within the same day, we drive to the suburban and inner-city areas. The weather was overcast and changed to rainy in the latter part of the recording. 
%%

%
\subsection{KITTI Drives}

\begin{figure}[!t]
    \centering
    \includegraphics[width=1.0\linewidth]{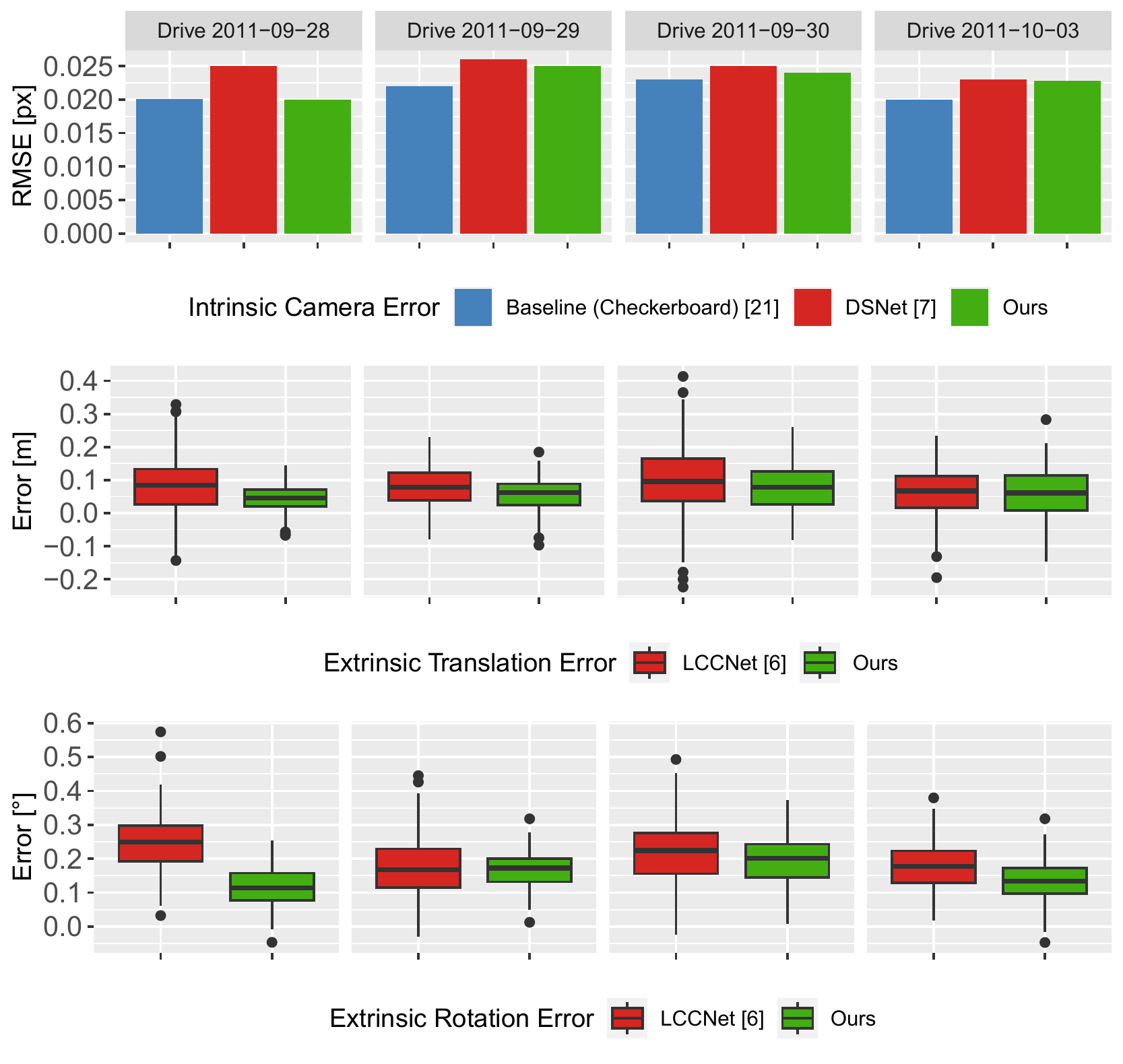}
    \caption[End-to-end Self-Calibration Pipeline]{Intrinsic and extrinsic calibration error. For intrinsic, the baseline is the best achievable RMS reprojection error obtained with the checkerboard-based calibration. For extrinsic, we calculate the mean error of all 3D axes. DSNet\cite{Zhang2020} and LCCNet\cite{Lv2021} were trained isolated and only single pass inference was used. See \cref{fig:vi_e2e_vs_iso} for the accompanying visualization and \cref{tab:calib_error} for mean values across all drives.}
    \label{fig:kitti_drive_error}
\end{figure}

The evaluation using KITTI drives is intended to show \gls{calica} accuracy and robustness when applied to diverse driving scenarios. Referring to \cref{fig:kitti_drive_error}, the mean error below 0.028 \si{px} across all four days of driving shows that our approach is comparable to classical checkerboard-based intrinsic calibration in terms of reprojection error. The significance of this low error can be visually inspected in \cref{fig:vi_e2e_vs_iso}. When the DSNet network is trained in isolation, we noted an error increase of almost 20\% relative to our proposed end-to-end approach (see drive 2011-09-28 in \cref{fig:kitti_drive_error}). Additionally, the undistortion result in \cref{fig:vi_e2e_vs_iso} (second image from the left) corroborates the increased error. Up to 15 pixels differences can be seen on the edge region of the image, where the distortion effect is more present. 

% \usepackage{booktabs}
\begin{table}[]
     \caption[Average Intrinsic and Extrinsic Calibration Errors]{Average Intrinsic and Extrinsic Calibration Errors} \label{tab:calib_error}
    \begin{tabular}{@{}lllll@{}}
        \toprule
        Calibration Error & Baseline\cite{opencv_library} & DSNet \cite{Zhang2020} & LCCNet\cite{Lv2021} & Ours  \\ \midrule
        Intrinsic [px]           & \textbf{0.018}                           & 0.024         & -             & 0.021 \\
        Extrinsic $\vb{t}$ [m]        & -                               & -             & 0.082         & \textbf{0.060} \\
        Extrinsic $\vb{R}$ [$\degree$]   & -                               & -             & 0.204         & \textbf{0.154} \\ \bottomrule
    \end{tabular}
\end{table}

A similar effect can be found in the extrinsic calibration: the end-to-end approach results in better Lidar-Camera alignment. Due to the intended use of online calibration, we put a constraint that the inference must happen in a single pass (no iteration). Under this constraint, we achieve a noticeably better mean rotational accuracy compared to when the LCCNet is trained in isolation (0.154$\degree$ vs. 0.203$\degree$).

We can infer some robustness by inspecting the statistical fluctuation of \gls{calica} prediction in \cref{fig:kitti_drive_error}. Across four days of driving with notably diverse sceneries (city, campus, residential, and road, see\cite{Geiger2013} for more details) the maximum intrinsic and extrinsic calibration errors remain below 0.028 \si{px} and 0.4$\degree$/0.2$\si{m}$, respectively. For extrinsic error, our outlier values are decisively lower than that of LCCNet, which we attribute to the Lidar-Camera measurement constraint realized by the weight sharing. Notwithstanding, we note the lowered accuracy in drive 2011-10-03 and bigger spread since the drive is dominated by highway sceneries (i.e., fewer edge and line-like features to extract). With this finding, we consider it beneficial if the car drives in a feature-rich environment (e.g., inner city) when the self-calibration functionality is active.

Overall, we consider the KITTI Drives evaluation result to appropriately reflects \gls{calica} fitness for vehicle sensors self-calibration. It fulfils its purpose of avoiding a miscalibration, as illustrated in \cref{fig:vi_e2e_vs_iso}; which can and will compromise the road object localization capability of a road vehicle. \newrobustcmd\B{\DeclareFontSeriesDefault[rm]{bf}{b}\bfseries}
\begin{table}[]
    \setlength{\tabcolsep}{1.0pt}
    \caption[Ablation study experiments]{Ablation Study Experiments} \label{tab:ablation}
    \begin{tabular}{@{}lllllllll@{}}
        \toprule
        Exp.\hspace{1pt}  & RGB train? & Depth train? & Siamese?  & $f$-loss  & $\xi$-loss  & $\vb{R}$-loss &  $\vb{t}$-loss &  \\ \midrule
        I & \ding{55} & \checkmark & \ding{55}   & -             & -          &$6.658\mathrm{E}\hspace{-0.75mm}-\hspace{-0.75mm}4\hspace{3mm}$
                 & 0.0014
           &             \\
        II &\checkmark & \ding{55} &\ding{55} & 11.41            & 0.070         & -         & -   &              \\
        II & \checkmark & \checkmark & \ding{55} & 11.41             & 0.068           & $6.054\mathrm{E}\hspace{-0.75mm}-\hspace{-0.75mm}4\hspace{3mm}$   & \textbf{0.0010 }   &              \\
        IV & \checkmark & \checkmark & \checkmark & \textbf{11.02}             & \textbf{0.067}           & ${4.758\mathrm{E}\hspace{-0.75mm}-\hspace{-0.75mm}4\hspace{3mm}}$    & \textbf{0.0010}    &              \\
        
        \bottomrule
    \end{tabular}

\end{table}
%\begin{table*}[]
%    \centering
%    \setlength{\tabcolsep}{2.5pt}
%    \caption[Comparison Result to Prior Works]{Comparison result to prior works} \label{tab:compare_sota}
%    \begin{tabular}{@{}llllllllll@{}}
%        \toprule
%        Method \#  & Miscalibration Range & Reproj. Error X & Trans. Mean X  & Trans. Mean Y   & Trans. MeanZ   & Rot. Mean X  &  Rot. Mean Y  &  Rot. Mean Z   \\ \midrule
%        DSnet (DeepPTZ) & 11.43 & 11.43 & 11.43   & 11.43             & 0.076           & 3.809e-4         & 0.0093   &   0.0093   &           \\
%        DSnet (Ours\cite{Rachman2022}) & 11.43 & 11.43 & 11.43   & 11.43             & 0.076           & 3.809e-4         & 0.0093   &    0.0093   &         \\
%        LCCNet [single-iteration] & 11.43 & 11.43 & 11.43 & 11.41             & 0.077           & 6.721e-4         & 0.0082    &     0.0093   &         \\
%        LCCNet [multi-iteration] & 11.43 & 11.43 & 11.43 & 11.41             & 0.077           & 6.721e-4         & 0.0082    &      0.0093   &        \\
%        Ours & 11.43 & 11.43 & 11.43 & 11.41             & 0.077           & 6.721e-4         & 0.0082    &      0.0093   &        \\
%  
%        \bottomrule
%    \end{tabular}
%    
%    
%\end{table*}

\subsection{Ablation Study}
We conducted ablation studies to evaluate the influence of end-to-end training on Lidar-Camera accuracy. Essentially, this study attempts to pinpoint which part of the pipeline is responsible for the improved performance. The evaluation metrics used are the validation terminal loss.

We used the identical setup in \cref{subsec:exp_setup}, but varied our training strategy and the corresponding network architecture. First, the feature extractors are no longer sharing weights. In experiment I, we froze the layers of the RGB extractor before training the network. In experiment II, we froze the layers of the depth extractor instead. In experiment III, all layers of the feature extractors are trainable, but the weight sharing is not activated. Finally, in Experiment IV, the normal end-to-end training with Siamese weight sharing was performed. The results can be seen in \cref{tab:ablation}.

It is quite evident that end-to-end training results in lower losses overall. The extrinsic parameters regressor (especially rotation with a decrease of 21\%) benefits the most when trained in both end-to-end and Siamese configurations (see $\vb{R}$-loss and $\vb{t}$-loss). However, we have observed that intrinsic calibration losses ($f$-loss and $\xi$-loss) do not appear to have a significant effect when trained end-to-end (loss decrease of less than 1\%). However, when the weight sharing is activated, we see a modest decrease in intrinsic losses (3\%). This may suggest that the feature-extracting networks cannot properly generalize the relevant feature for camera intrinsic parameters.

The behaviour of the intrinsic calibration network can be potentially explained by insufficient edge coverage from the KITTI checkerboard recording. The distortion on the far-edge region of the image is not fully modelled and may result in suboptimal baseline calibration $\theta_{0}$, and by extension, $\theta_{label}$. In addition, the USM model used does not consider the tangential and radial distortion\cite{Mei2007}. Based on this study, we learned the limitation of our label generation strategy, which heavily relies on good checkerboard recording. For real applications and future works, the training label should be based on a checkerboard recording of sufficient coverage and tangential and distortion models should be considered.

%
%
%
%In the first part, we compare calibration accuracy when the calibration is done in isolation. First, the intrinsic  calibration is done with DSNet. The resulting intrinsic parameter is used as an input to undistort images which are one of the input of LCCNet. LCCNet then predicts the Lidar-camera extrinsic parameter. 
%
\section{Conclusion}\label{AA}
We have proposed a novel end-to-end learning strategy for jointly self-calibrating camera and Lidar sensor with the accompanying deep neural network \gls{calica}. The pipeline proposed a label generation strategy with a normally operating vehicle and existing calibration infrastructure. It also ensures the quality of the inputs to the self-calibration pipeline by means of detecting SIFT key-points and detecting Lidar-Camera measurements overlap. Our main contribution lies in the design of \gls{calica}, consisting of two feature extractor networks (one for camera image and one for Lidar point cloud) arranged in a Siamese structure. The design is devised to deal with the limitation of using single sensor modality (i.e., bottlenecked by a single sensor physical limitation) and mitigating the propagation of error when estimating extrinsic calibration using unverified intrinsic calibration.

Finally, we performed thorough verification and validation based on an ablation study and evaluation with realistic driving scenarios. Our approach is shown to perform well when compared to classical, infrastructure-bounded checkerboard calibration. Furthermore, compared to other learning-based approaches, our approach can outperform isolated intrinsic and extrinsic self-calibration with a single-pass inference, paving its way to adoption in real-world online applications.

\bibliographystyle{IEEETran}
\bibliography{ref_corrected}

\end{document}